\documentclass[letterpaper, 10 pt, conference]{IEEE/ieeeconf}  
\usepackage[utf8]{inputenc}
\usepackage{algorithm,algorithmic}
\usepackage{graphicx}
\usepackage{amsmath}
\usepackage{xcolor}
\usepackage{hyperref}

\IEEEoverridecommandlockouts                              

\usepackage{graphics} 
\usepackage{epsfig} 
\usepackage{mathptmx} 
\usepackage{times} 
\usepackage{amsmath} 
\usepackage{amssymb}  
\let\labelindent\relax
\usepackage[inline]{enumitem}
\usepackage{graphicx}
\usepackage{subfig}
\usepackage{float}
\newcommand\norm[1]{\left\lVert#1\right\rVert}
\DeclareMathOperator*{\argmin}{argmin} 

\title{\LARGE \bf
Look Both Ways: Bidirectional Visual Sensing for Automatic Multi-Camera Registration
}
\date{February 2020}

\author{Subodh Mishra$^{*+}$, Sushruth Nagesh$^{+}$, Sagar Manglani, Graham Mills$^{**}$, \\Punarjay Chakravarty and Gaurav Pandey
\thanks{$^{*}$S. Mishra (subodh514@tamu.edu) is with Texas A$\&$M University (work done as intern), others are with Ford Greenfield Labs, Palo Alto and can be reached at snagesh1@ford.com, smanglan@ford.com, pchakra5@ford.com, gpandey2@ford.com}%
\thanks{$^{+}$ S. Mishra and S. Nagesh have equally contributed to the work.}
\thanks{$^{**}$This work was performed by G. Mills (grayhem@gmail.com) when they were employed at Ford.}
}
\begin{document}
\maketitle
\thispagestyle{empty}
\pagestyle{empty}

\begin{abstract}
This work describes the automatic registration of a large network $(\approx 40)$ of fixed, ceiling-mounted environment cameras spread over a large area $(\approx 800$ $m^{2})$ using a mobile calibration robot equipped with a single upward-facing fisheye camera and a backlit ArUco marker for easy detection. The fisheye camera is used to do visual odometry (VO), and the ArUco marker facilitates easy detection of the calibration robot in the environment cameras. In addition, the fisheye camera is also able to detect the environment cameras. This two-way, bidirectional detection constrains the pose of the environment cameras to solve an optimization problem. Such an approach can be used to automatically register a large-scale multi-camera system used for surveillance, automated parking, or robotic applications.  This VO based multi-camera registration method has been extensively validated using real-world experiments, and also compared against a similar approach which uses a LiDAR - an expensive, heavier and power hungry sensor.
\end{abstract}
\section{INTRODUCTION}
\begin{figure}[]
\centering
\includegraphics[width=0.45\textwidth]{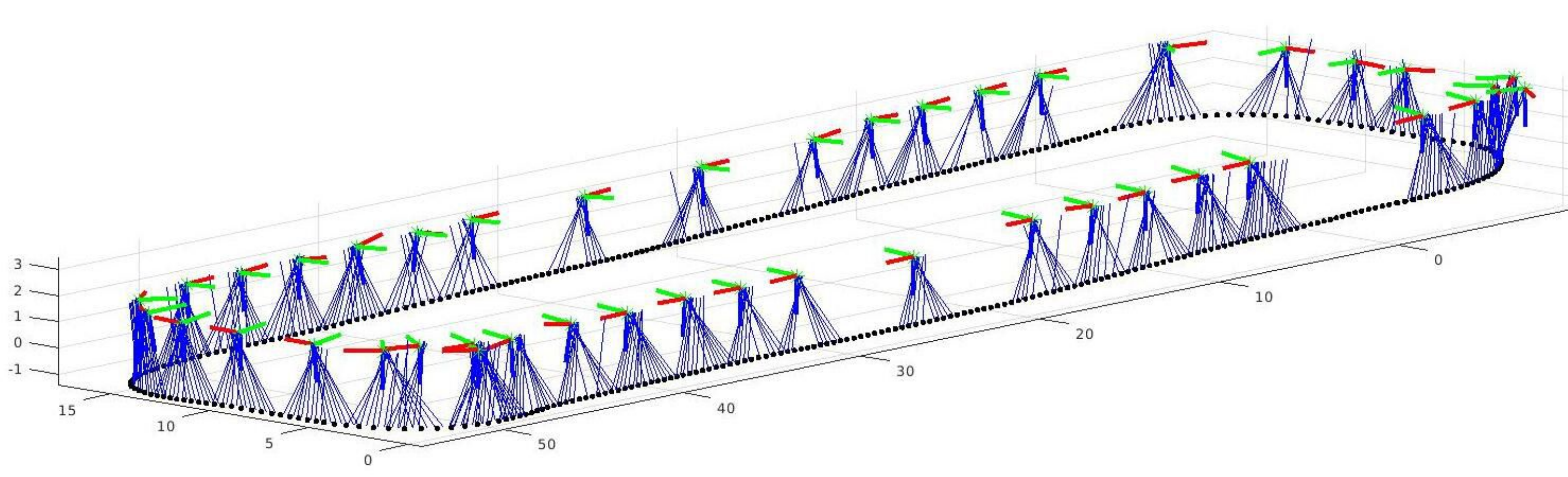}
\caption{\small{Plot of estimated poses of multiple cameras (shown as axes) and calibration robot trajectory (dark dots). The blue straight lines are fiducial marker detections from the environment cameras.}}
\label{fig: final-result}
\vspace{-2.25em}
\end{figure}

\begin{figure}[h]
\centering
\includegraphics[width=0.45\textwidth]{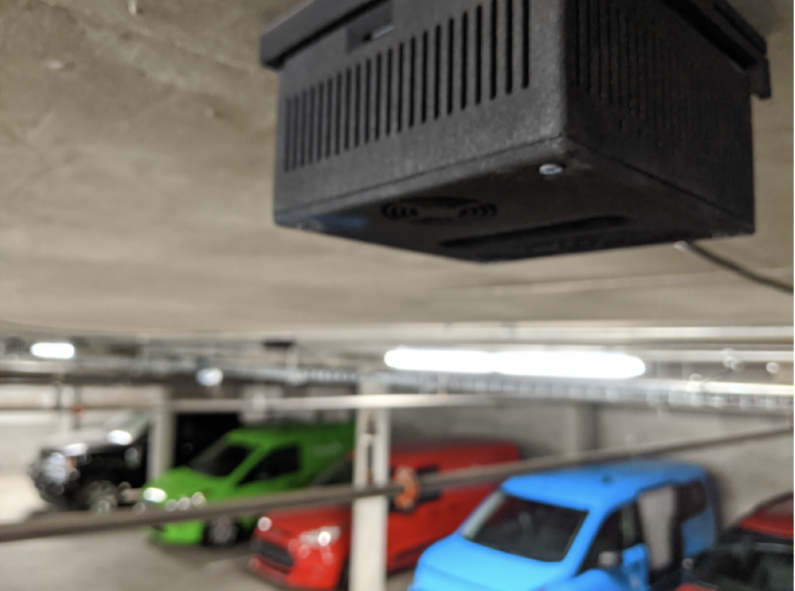}
\caption{\small{\textbf{Edge nodes:} We register a distributed sensing system of $\sim$ 40 edge devices mounted throughout the ceiling of our test environment. Each edge node comprises of an RGB camera (we call it as environment camera in the paper) and compute. These are powered over ethernet and networked and time-synced to a central server.}}
\label{fig: edge-node}
\vspace{-2.15em}
\end{figure}

Infrastructure Enabled Autonomy \cite{IEA} is gaining popularity with the aim to equip the environment of an autonomous system with sensors that can eliminate the dependence on onboard sensors for perception and localization, especially in controlled environments like parking lots, warehouses, and factory floors. A distributed sensing system in a parking lot can autonomously park vehicles with drive-by-wire technology and direct the movement of cars through the environment. However, for such a network of fixed environment cameras to be useful to an autonomous robot for the aforementioned applications, they need to be registered w.r.t. each other such that the 6 DoF pose of each environment camera is known w.r.t. other environment cameras.  A tedious way to determine the camera poses would involve detecting a calibration target (ArUco/Checkerboard) \cite{GarridoJurado2014AutomaticGA} in the common Field of View (FoV) between every pair of environment cameras, determining their relative poses and subsequently daisy-chaining the poses through the sequence of cameras and back to the starting camera. This approach not only needs significant overlapping field of view (FoV) that cannot always be guaranteed in many practical scenarios, but also tiresome manual labor when a large number of cameras need to registered over a large area.  
\vspace{-1em}
\section{RELATED WORK}
One of the first works to describe the joint calibration of a system of surveillance cameras and a mobile robot \cite{chakravarty09acra} used homography between the image plane of each camera and the ground plane to calibrate the two. As the robot moved around the FoV of each camera, its simultaneous detection in the camera image plane and localization using LiDAR on the ground plane allowed the automated computation of homography between each camera and the ground plane. More recently, Zou et. al. \cite{Zou22icra} describe a set of fisheye cameras deployed at an intersection to monitor traffic. Each camera-node (comprising of sensor and compute) is extrinsically calibrated using homography between hand-picked points on the road (visible in the camera image) that are also discernible in a satellite image of the area. \cite{Mishra22icra} describes a method to register a single fisheye camera in a given map, but, as the title suggests the method requires the foreknowledge of prior mapping data. Registration of multiple cameras with over-lapping fields of view is an easier problem to solve, for example by using Epipolar geometry. A comprehensive account of challenges in registration of multiple cameras with no or minimal common field of view is presented in \cite{XIA2018951}. \cite{BridgingTheGapBetweenCameras} $\&$ \cite{SimCalAndTrackNetwork} present methods of camera registration in which static non overlapping cameras track pedestrians and in the process determine the registration between 4 $\&$ 6 cameras respectively spread over a small area. Similarly, \cite{CameraLocalizationinDistributedNetworksUsingTrajectoryEstimation} presents a Kalman Filter based method to register 4 static cameras in the same plane by using trajectory estimation of a calibration target in an area of $\approx$ 16 $m^{2}$. \cite{BridgingTheGapBetweenCameras}, \cite{SimCalAndTrackNetwork} and \cite{CameraLocalizationinDistributedNetworksUsingTrajectoryEstimation} are similar in their approach of tracking an object to determine metric topology of a network of cameras but these experiments have been carried out for small number of cameras and in a small area. \cite{CalibrationofNonoverlappingCamerasUsinganExternalSLAMSystem} uses a RGB-D SLAM \cite{PointplaneSLAMforhandheld3Dsensors} to generate a prior map of 3D-points with visual intensity information, and then performs registration of 3 static non over-lapping cameras in it by 2D-3D correspondences followed by solving a PnP problem. \cite{Calibrationofnonoverlappingcamerasbasedonamobilerobot} presents a method to register only two non over-lapping cameras that are less than a meter apart and rely heavily on the odometry of the calibration robot which comes from a wheel encoder which is prone to drift. \cite{Markerbasedsimplenonoverlappingcameracalibration} presents a technique where they register 2 static non over-lapping cameras. In this case the cameras to be registered need to have fiducial markers on them. The system requires a support calibration camera that can view the fiducial markers on the cameras to be registered  and two different checkerboard targets simultaneously. This system will yield precise registration results for small system of cameras but has practical limitations on extending it to a system of several cameras spread over a large area. For the scenario presented in \cite{Markerbasedsimplenonoverlappingcameracalibration}, registering just 2 cameras not only required 1 support camera but also 4 fiducial markers in total, making this system impractical for large scale large area registration. Unlike \cite{CalibrationofNonoverlappingCamerasUsinganExternalSLAMSystem} that registers cameras in a prior RGB-D map, \cite{AvisualSLAMbasedapproachforcalibrationofdistributedcameranetworks} registers 6 static non over-lapping cameras using a sparse 3D map obtained from a Visual SLAM \cite{ORBSLAM} pipe-line by using selected manual 2D-2D correspondences between the SLAM camera and the environment camera. The limitation of \cite{AvisualSLAMbasedapproachforcalibrationofdistributedcameranetworks} is that it is not in metric scale, it requires manual selection of features and the mapping camera and the environment camera must capture images of the same feature rich areas - which is not possible in our scenario where the cameras to be registered look vertically below on a largely texture less surface which is not suitable for doing visual odometry/SLAM. \cite{NonoverlappingRGBDCameraNetworkCalibrationwithMonocularVisualOdometry} presents a technique to register 5 static RGB-D cameras out of which some share field of view and some do not. The process bridges camera poses by visual odometry in which the visual odometry camera and the RGB-D cameras need to co detect several fiducial markers in the scene. Finally, a depth based refinement process is undertaken using the RGB-D cameras. The limitation of this process is that it requires almost as many fiducial markers as the number of RGB-D cameras to be calibrated and these fiducial markers should also be co-detected in the visual odometry and the cameras being registered. This is practically not possible in scenarios involving a large number of cameras spread over a large area.

\section{CONTRIBUTIONS}
\label{sec: contributions}
In this work, we present an approach to automatically register a large network $(\approx 40)$ of static ceiling-mounted RGB cameras (Figures \ref{fig: edge-node} $\&$ \ref{fig: distributed-cameras}) with minimal or no overlap between them, spread over a large area $(\approx 800$ $m^{2})$ using a calibration robot (Figure \ref{fig: calibration-robot}) equipped with a single upward looking fisheye camera - for visual odometry (VO) \cite{visodomtutorial1} $\&$ detection of environment cameras (Figure \ref{fig: distributed-cameras}), and a single back-lit ArUco \cite{GarridoJurado2014AutomaticGA} marker - for easy detection of calibration robot in environment cameras (Figure \ref{fig: view-from-distributed-camera}). We use VO and ArUco marker detection to bridge the pose between environment cameras. We look/sense in both directions: downwards from each of the environment camera down onto the robot (as it passes through its FoV, Figure \ref{fig: view-from-distributed-camera}) and, upwards from the robot to the environment camera (Figure \ref{fig: distributed-cameras}) to determine the pose of  each environment camera relative to a common coordinate frame.
\section{SYSTEM DESCRIPTION $\&$ NOTATION}
\subsection{The Distributed Camera System} 

\begin{figure}[h]
\centering
\includegraphics[width=0.45\textwidth]{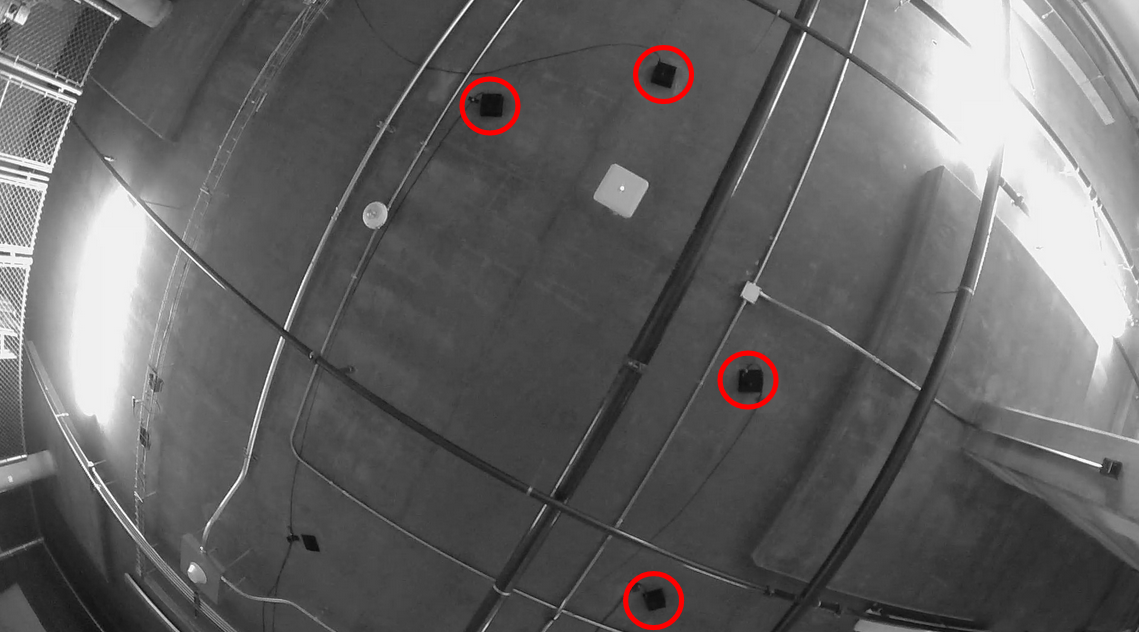}
\caption{\small{\textbf{Looking up $\uparrow$ Edge-Node detection from Robot:} Automated detection of ceiling-mounted environment cameras when viewed from the upward-facing fisheye camera on the calibration robot.}}
\label{fig: distributed-cameras}
\vspace{-1.5em}
\end{figure}
The distributed camera system (Figures \ref{fig: edge-node} $\&$ \ref{fig: distributed-cameras}) comprises several static environment cameras which have minimal to no overlapping FoV. They generate images of size 1080 x 1920 pixels at 30 Hz. We use the factory provided intrinsic calibration parameters for these cameras.

\subsection{The Calibration Robot}
\begin{figure}[h]
\centering
\includegraphics[width=0.45\textwidth]{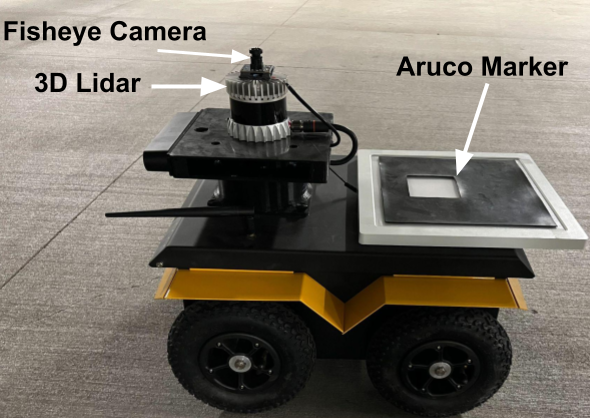}
\caption{\small{\textbf{Calibration Robot:} Clearpath Jackal with an upward-facing wide-angle fisheye camera $F$ and an ArUco marker $M$. The ArUco marker is backlit to make detection easier. The robot also carries a LiDAR for comparison between LiDAR and camera-only calibration.}}
\label{fig: calibration-robot}
\vspace{-1.25em}
\end{figure}

\begin{figure}[H]
\centering
\includegraphics[width=0.45\textwidth]{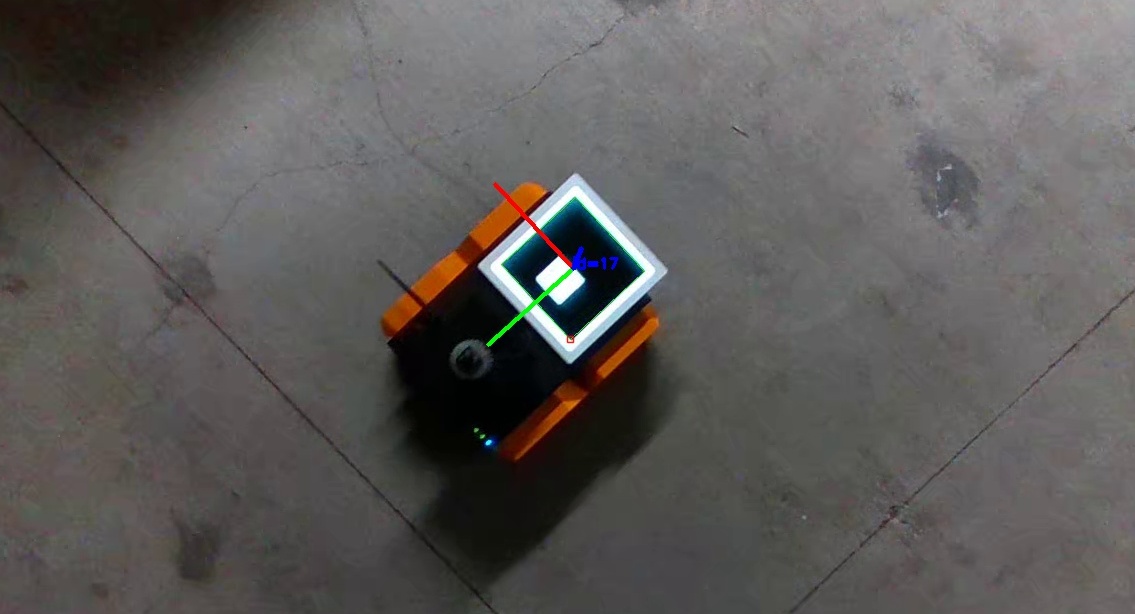}
\caption{\small{\textbf{Looking down $\downarrow$ Robot Detection from Edge Node:} The Calibration Robot (Figure \ref{fig: calibration-robot}) viewed from an environment camera with axes drawn on the detected backlit ArUco marker.}}
\label{fig: view-from-distributed-camera}
\vspace{-1.25em}
\end{figure}

We use a Clearpath Jackal as our calibration robot (Figure \ref{fig: calibration-robot}). The robot has a rigidly attached upward-facing wide-angle (160$^{\circ}$) 2MP fisheye camera which generates images of size 1080 x 1920 pixels at 30 Hz. The fisheye camera faces upwards because the ceiling (height varying between 3 - 4 m from ground level) of our indoor facility has sufficient static features to track for robot pose estimation using VO. We use a wide-angle lens for VO such that we can cover a large FoV and track several features. Moreover, a wide-angle camera ensures that several environment cameras can be detected (using OpenCV \cite{opencv_library} blob detection, Figure \ref{fig: distributed-cameras}), which we use to further constrain the position of the detected cameras in the estimation procedure. We use the fisheye camera calibration toolbox available in MATLAB \cite{ocamcalib} to obtain its intrinsics. A backlit ArUco marker \cite{GarridoJurado2014AutomaticGA} of known dimensions is attached to the calibration robot so that the robot can be easily detected by the environment cameras (Figure \ref{fig: view-from-distributed-camera}).  The calibration robot also has a 64 Channel Ouster 3D-LiDAR which generates 3D scans at 10 Hz used for doing LiDAR Odometry (LO) \cite{DLO} based calibration of environment cameras - a method we will compare our approach against.
\subsection{Notation}
In this work, 6-DoF pose is expressed as a homogeneous transformation matrix $T = \begin{bmatrix}
R(q) & p\\
0 & 1
\end{bmatrix}$, where rotation $R$ is represented as quaternion $q$ and position $p$ as a $R^{3\times 1}$ vector. The global frame $W$ coincides with the origin of the robot motion estimated using VO. The pose $T^{W}_{F}$ of the fisheye camera $F$ is synonymous with the calibration robot pose. We have $N$ cameras $\{C_k\}_{k=0:N-1}$ in the distributed environment camera system with corresponding global poses $\{T^{W}_{C_k} \}_{k=0:N-1}$.  The detection of the ArUco marker $M$ on the calibration robot in an environment camera $C_k$ results in ArUco detection pose measurement $T^{C_k}_{M}$ which represents the pose of the ArUco marker in camera ${C_k}$ frame. The fisheye camera $F$ and the ArUco marker $M$ have a fixed spatial offset $T^{F}_{M}$ between them. The fisheye camera $F$ detects and tracks point features (as pixels $\in R^{2}$) to do VO for robot pose estimation. The fisheye camera also detects environment cameras $\{C_{k}\}_{k=0:N-1}$ as 2-D pixel ($\in R^{2}$) measurements on its image plane which are ultimately used to constrain the position of the environment camera. 

\subsection{Time Synchronization}
We synchronize all edge nodes (environment cameras and calibration robot) using an NTP (Network Time Protocol) server. A central server acts as a NTP server and all the edge nodes use this server to sync their time as NTP clients. The resulting synchronization has a time error in the range of 1 to 2 ms on the client devices w.r.t the time on the server. Also, the sensors (cameras and LiDAR on the robot vs the environment cameras) are asynchronous (not triggered together, different frame rates) leading to a small variable time error between them. This error will have negligible impact on registration because we operate the calibration robot (Figure \ref{fig: calibration-robot}) at speeds significantly lower than its top speed ( which is 2 m/s). 

\section{PROBLEM STATEMENT}
The primary goal of this work is to determine the pose $\{ T^{W}_{C_k}\}_{k=0:N-1} \in SE(3)$ of environment cameras $\{C_k\}_{k=0:N-1}$ (Figure \ref{fig: distributed-cameras}) with respect to a global frame $W$. As the environment cameras share no or minimal FoV, the calibration robot's fisheye camera pose $T^{W}_{F} \in SE(3)$ and the constant spatial offset $T^{F}_{M} \in SE(3)$ between the fisheye camera $F$ and the ArUco marker $M$ also need to be estimated such that an ArUco detection pose measurement like $T^{C_{k}}_{M} \in SE(3)$ can be used to estimate $T^{W}_{C_k}$ - camera $C_k$'s global pose (since $T^{W}_{C_{k}} = T^{W}_{F} T^{F}_{M} (T^{C_k}_{M})^{-1}$). In addition, the scale $s$ of robot motion also needs to be estimated because the robot pose $T^{W}_{F}$ estimated from VO is not in metric scale. 

\section{METHOD}
We follow a 3 step approach to solve this problem. They are: 
\begin{enumerate*}[label=(\Alph*)]
  \item Estimate unscaled robot motion $T^{W}_{F}$ (and unscaled map $\{X_{k}\in R^{3}\}_{k=0:P-1}$ of 3D points from 2D feature detection and tracking) with the upward-looking fisheye camera $F$ using Visual Odometry (VO) (Section \ref{sec: visual-odometry}),
  \item Estimate the scale $s$ of VO and the spatial offset $T^{F}_{M}$ between the upward-looking fisheye camera $F$ and the ArUco marker $M$ using detection of ArUco marker in the environment cameras (Section \ref{sec: mocal}),
  \item Estimate the poses $\{ T^{W}_{C_k}\}_{k=0:N-1}$ of the distributed environment cameras $\{C_k\}_{k=0:N-1}$ using two way measurements i.e. ArUco detection pose measurements from detection of ArUco marker in the environment cameras $\{C_k\}_{k=0:N-1}$ and their detection in the fisheye camera $F$ (Section \ref{sec: estimateenvironmentcamerapose}).
\end{enumerate*}

\subsection{Estimate Motion of the upward-looking fisheye camera using Visual Odometry (VO)}
\label{sec: visual-odometry}
\begin{figure}[h]
\centering
\includegraphics[width=0.495\textwidth]{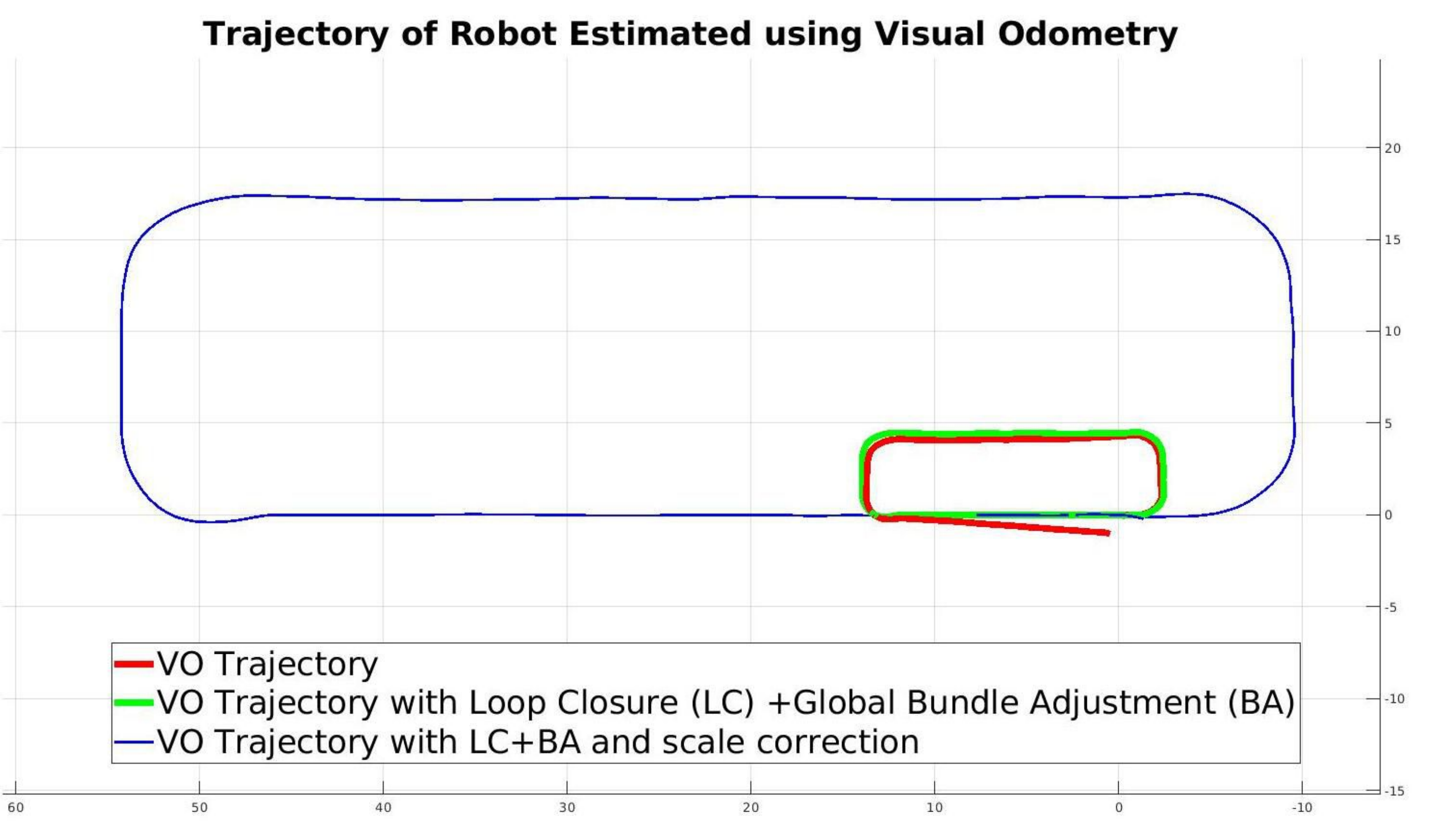}
\caption{\small{\textbf{Motion estimation using Visual Odometry (VO).} The trajectory in red is the result from SVO \cite{Forster14icra}'s front-end. We use loop closure \cite{GalvezTRO12} and bundle adjustment \cite{ceres-solver} to reduce the drift in VO (shown in green) and finally we determine the metric scale (in meters) (Section \ref{sec: mocal}) of the trajectory using the detection of ArUco marker on the calibration robot.}}
\label{fig: vo-compare-graph}
\vspace{-1em}
\end{figure}

We use SVO \cite{Forster14icra}'s front-end to determine the frame to frame motion of the upward-looking fisheye camera on the robot. We choose SVO over other VO algorithms because of its ease of use and proven ability with wide-angle lenses \cite{Zhang16icra}. Additionally, SVO can be optimized to work for downward-looking cameras in aerial robots where the camera tracks features on a dominantly flat surface, a situation which is similar to our case where we use an upward-looking camera that tracks static features on the ceiling. We eliminate drift in VO by implementing loop closure (using DBoW2 \cite{GalvezTRO12} for place recognition) and performing bundle adjustment \cite{BAinthelarge} (using ceres solver \cite{ceres-solver}). A comparison of trajectories before and after bundle adjustment is shown in Figure \ref{fig: vo-compare-graph} (compare red and green).

Because of the asynchronous nature of data generated from several sensors we use pose interpolation to associate a robot pose to each ArUco detection. If the ArUco detection has a timestamp $t \in (t_i, t_j)$ where $t_i$ and $t_j$ are the timestamps of the robot keyframes $F_i$ and $F_j$ with poses $T^{W}_{F_i} = \begin{bmatrix}
R(q^{W}_{F_i}) & p^{W}_{F_i}\\
0 & 1
\end{bmatrix}$ and $T^{W}_{F_j} = \begin{bmatrix}
R(q^{W}_{F_j}) & p^{W}_{F_j}\\
0 & 1
\end{bmatrix}$ respectively, then the robot pose $T^{W}_{F}(t)$ at time $t$ is determined by taking weighted average of $p^{W}_{F_i}$ $\&$ $p^{W}_{F_j}$ - for the translation component, and \texttt{slerp} (Spherical Linear Interpolation \cite{slerppaper}) - for the rotation component.
This method allows us to associate each ArUco detection $T^{C_k}_{M}(t)$ with an interpolated robot pose $T^{W}_{F}(t) \in SE(3)$.
\subsection{Determine the scale $s$ of VO $\&$ the spatial offset $T^{F}_{M}$ between the ArUco marker $M$ and the upward-looking fisheye camera $F$ using ArUco detections}
\label{sec: mocal}
We use the ArUco detection pose measurements $T^{C_k}_{M} \in SE(3)$ (this is the pose of the marker $M$ in frame $C_k$) to determine the metric scale $s$ of the trajectory estimated by monocular VO and the spatial separation $T^{F}_{M}$ between the fisheye camera and the ArUco marker. The motion-based calibration technique \cite{MocalZachTaylor} is employed for this purpose, using which we align the motion estimated by VO with the motion of the ArUco marker estimated by its detection in an environment camera $C_k$. For two adjacent robot keyframes $F_i$ $\&$ $F_{i+1}$, we search for the nearest ArUco detection measurements $T^{C_k}_{M_i}$ $\&$ $T^{C_k}_{M_{i+1}}$ in the same environment camera $C_k$ by comparing timestamps. Next, we use pose interpolation (discussed before) to determine the robot pose ${}_{int}T^{W}_{F_i}$ $\&$ ${}_{int}T^{W}_{F_{i+1}}$ corresponding to timestamps of ArUco measurements $T^{C_k}_{M_i}$ $\&$ $T^{C_k}_{M_{i+1}}$ respectively. Our goal is to align $_{int}T^{F_i}_{F_{i+1}}$ $(= ({}_{int}T^{W}_{F_i})^{-1}{}_{int}T^{W}_{F_{i+1}})$ with $T^{M_i}_{M_{i+1}}$ $(= (T^{C_k}_{M_i})^{-1} T^{C_k}_{M_{i+1}})$ and in the process determine the metric scale $s$ of robot trajectory and also the spatial offset $T^{F}_{M}$ between the fisheye camera $F$ and the marker $M$. We model the VO trajectory with a scale factor $s$ multiplied to the translation component ${}_{int}p^{F_i}_{F_{i+1}}$.  We solve a non-linear least squares optimization problem to determine the unknown variables. The residual $r_{i}$ for this optimization process is given by:
\begin{equation}
    r_{i} = {}_{int}T^{F_{i}}_{F_{i+1}} T^{F}_{M} \ominus T^{F}_{M} T^{M_{i}}_{M_{i+1}}
    \label{eqn: mocal}
\end{equation}
The symbol $\ominus$ differentiates the operation from a standard subtraction in Euclidean space and is meant to demonstrate operations on manifolds \footnote{\href{https://rss2017.lids.mit.edu/docs/invitedtalks/park-manifoldsgeometryandrobotics.pdf}{https://rss2017.lids.mit.edu/docs/invitedtalks/park-manifoldsgeometryandrobotics.pdf}}. We model ${}_{int}T^{F_{i}}_{F_{i+1}} = \begin{bmatrix}
R({}_{int}q^{F_{i}}_{F_{i+1}}) & s . {}_{int}p^{F_{i}}_{F_{i+1}}\\
0 & 1
\end{bmatrix}$. We use the ceres solver \cite{ceres-solver} to solve for $s$ and $T^{F}_{M}$ by minimizing a cost function shown in Equation \ref{eqn: mocal-optimization}.
\begin{equation}
    \hat{s}, \hat{T}^{F}_{M} = \argmin_{s\in R, T^{F}_{M} \in SE(3)} \sum_{i=0}^{L-1} \rho(\norm{r_{i}}^{2})
    \label{eqn: mocal-optimization}
\end{equation}
Here $L$ is the number of corresponding motion segments and $\rho()$ is Huber Loss Function. The estimated scale $s$ and spatial offset $T^{F}_{M}$ are used to re-scale VO pose estimates (Figure \ref{fig: vo-compare-graph}) and to estimate the environment camera poses (Section \ref{sec: estimateenvironmentcamerapose}). 
\subsection{Estimate the poses of the environment cameras}
\label{sec: estimateenvironmentcamerapose}
\subsubsection{Looking Down} 
\label{sec: estimateposes-lookingdown}
We use the ArUco pose measurements from each environment camera of the distributed system to determine its pose in a global coordinate system. For a single camera $C_k$ of the distributed system, we gather all the ArUco pose measurements $\{T^{C_k}_{M_i}\}_{i=0:N_{k}-1}$ which measure the poses of the ArUco marker in the camera $C_k$ frame as the calibration robot drives under it, and use their respective timestamps to determine the corresponding robot/fisheye camera pose $\{_{int}T^{W}_{F_i}\}_{i=0:N_{k-1}}$ using interpolation. For ArUco measurements $\{T^{C_k}_{M_i}\}_{i=0:N_{k}-1}$ from camera $C_k$, and the corresponding interpolated robot poses $\{{}_{int}T^{W}_{F_i}\}_{i=0:N_{k-1}}$, the pose of environment camera $C_k$ can be determined from Equation \ref{eqn: initial-pose} by solving a non linear least squares optimization problem using ceres library \cite{ceres-solver}.
\begin{align}
    \hat{T}^{W}_{C_{k}} = \argmin_{T^{W}_{C_{k}} \in SE(3)} \sum_{i=0}^{N_k-1} \rho(\norm{T^{W}_{C_{k}} T^{C_k}_{M_i} \ominus {}_{int}T^{W}_{F_i}\hat{T}^{F}_{M}}^{2})
    \label{eqn: initial-pose}
\end{align}
Here $N_k$ is the number of ArUco detection pose measurements made by camera $C_k$ and $\rho()$ is Huber Loss Function. We perform this optimization (Equation \ref{eqn: initial-pose}) for all environment cameras $\{C_k\}_{k=1:N}$ to estimate their respective 6-DoF pose using respective ArUco detection pose measurements made by looking down on the calibration robot driving below. 
\subsubsection{Looking Up - Position Refinement}
\label{sec: estimateposes-lookingup}
In this step our goal is to refine the positions $\{ p^{W}_{C_k}\}_{k=0:N-1} \in R^{3}$ of the estimated camera poses $\{ T^{W}_{C_k}\}_{k=0:N-1} \in SE(3)$ by minimizing the re-projection error between the projection $\pi (K, T^{W}_{F_{_j}}, p^{W}_{C_k})$ of the estimated environment camera position $p^{W}_{C_k}$ and the corresponding pixel detection $\begin{bmatrix}
u \\ v
\end{bmatrix}_{kj}$ (obtained using OpenCV's blob detection algorithm) on the upward facing fisheye camera image. We associate $\pi (K, T^{W}_{F_{_j}}, p^{W}_{C_k})$ to $\begin{bmatrix}
u \\ v
\end{bmatrix}_{kj}$ by performing a nearest neighbour search. The residual for minimizing the reprojection error is given in Equation \ref{eqn: ceiling-camera-re-projection-residual}.
\begin{align}
r_{kj} &= \begin{bmatrix}
u \\ v
\end{bmatrix}_{kj} - \pi (K, T^{W}_{F_{_j}}, p^{W}_{C_k})
\label{eqn: ceiling-camera-re-projection-residual}
\end{align}
$r_{kj}$ can be defined as the reprojection error of the $k^{th}$ environment camera when viewed from the $j^{th}$ robot key-frame pose. Here $K$ $\&$ $\pi$ are the intrinsic calibration parameters of the fisheye camera and the fisheye projection model respectively. We solve this minimization problem (Equation \ref{eqn: ceiling-camera-pose-optimization}) using ceres library \cite{ceres-solver}.
\begin{align}
    \{\hat{p}^{W}_{C_k}\}_{k=0:N-1} &= \argmin_{\{p^{W}_{C_k}\}_{k=0:N-1}\in R^{3\times 1}} \sum_{j=0}^{M-1} \sum_{k=0}^{N-1} w_{kj} \rho(\norm{r_{kj}}^{2})
    \label{eqn: ceiling-camera-pose-optimization}
\end{align}
Here $w_{kj} = 1$ if environment camera $C_k$ is visible in the $j^{th}$ robot key-frame otherwise $w_{kj} = 0$ and $\rho()$ is Cauchy Loss Function.  Figure \ref{fig: final-result} shows a plot of estimated environment cameras, ArUco detection pose measurements and the calibration robot trajectory after the registration procedure. Equation \ref{eqn: ceiling-camera-pose-optimization} is similar to solving a Bundle Adjustment problem.

\section{EXPERIMENTS \& RESULTS}
\begin{figure}[h]
        \centering
        \includegraphics[width=0.45\textwidth]{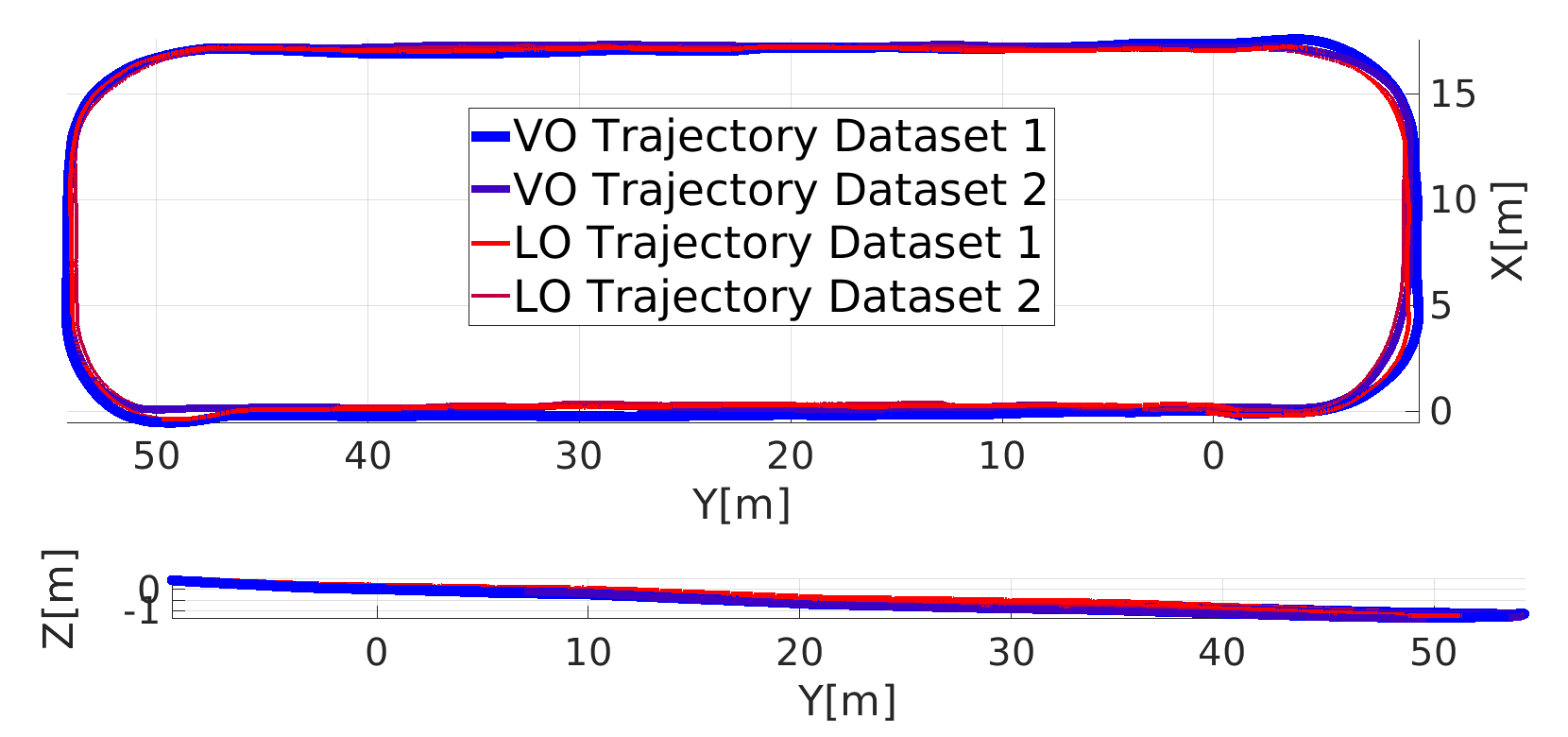}
        \label{fig: placeholder1}
        \caption{\small{Calibration Robot Trajectory as estimated by Visual Odometry and LiDAR odometry. We drive the calibration robot under the loop of environment cameras that need to be registered. [The trajectories significantly overlap, making it difficult to discern four different trajectories here.]}}
        \vspace{-1.5em}
\end{figure}

To evaluate our method, we collect two datasets by driving the calibration robot under the environment cameras of the distributed system. We register 43 and 38 environment cameras in datasets 1 $\&$ 2 respectively. We also collect 3D scans from the LiDAR on the calibration robot for performing qualitative comparison of the proposed VO based solution against a LiDAR Odometry LO \cite{DLO} based approach. One of the immediate advantages of the VO based method over the LO based approach is that one can perform qualitative verification of environment camera registration by projecting environment cameras' positions on the fisheye camera image.
\vspace{-1.5em}
\subsection{Verification using Re-projection Error}
\label{sec: self-contained-verification}
\begin{figure*}[]
        \centering
        \subfloat[\textbf{Re-projection before camera position refinement.} The detected environment cameras (green) do not align with projection of environment cameras (red). ]{\includegraphics[width=.45\textwidth]{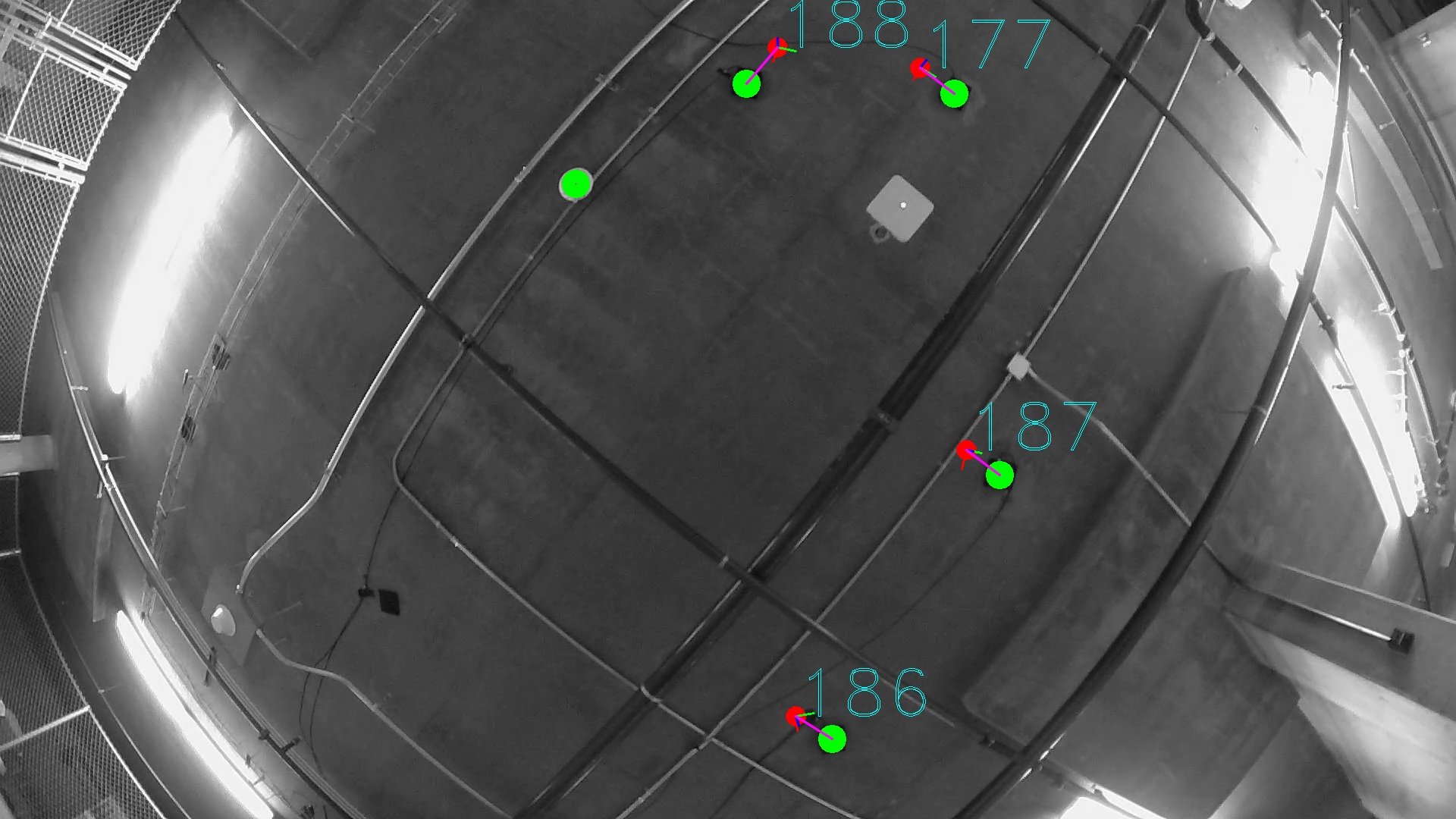}\label{fig: visual-residual}}
        \quad 
        \subfloat[\textbf{Re-projection after camera position refinement.} The projection of estimated environment cameras coincide with corresponding environment cameras on the fisheye image.]{\includegraphics[width=.45\textwidth]{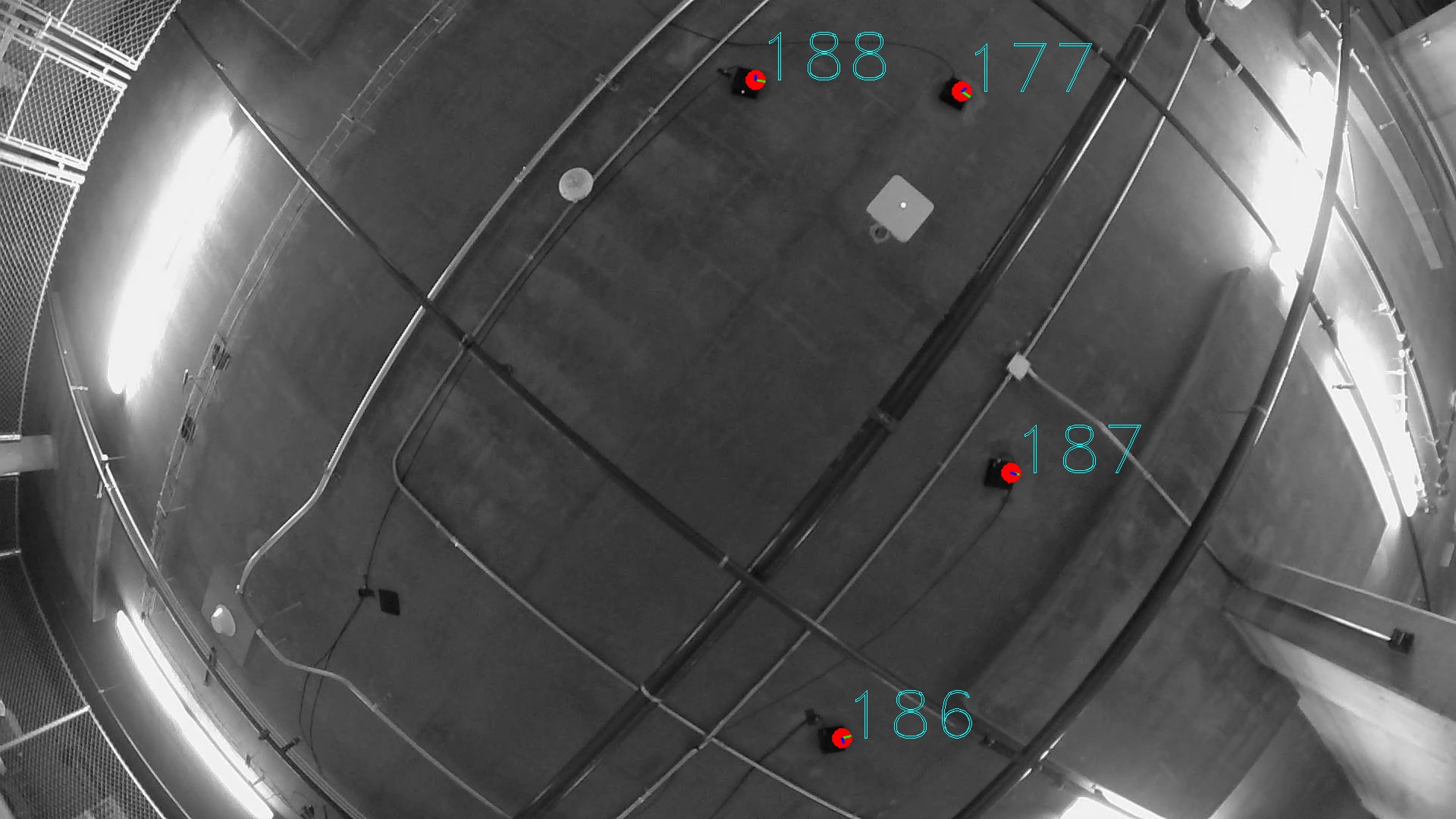}\label{fig: visual-residual-optimized}} 
        \caption{\small{Re-projection of distributed camera positions before and after re-projection error based refinement of camera positions.}}
        \vspace{-1.25em}
\end{figure*}

We present the reduction in environment camera re-projection errors before and after refinement (Section \ref{sec: estimateposes-lookingup}) of estimated environment camera positions in Figures \ref{fig: visual-residual} $\&$ \ref{fig: visual-residual-optimized} and tabulate the result in Table \ref{table: reprojection-error}.
\begin{table}[h]
\centering
\resizebox{0.75\columnwidth}{!}{%
\begin{tabular}{| c |  c  | c |} 
 \hline
 \multicolumn{3}{|c|}{Reprojection Error (Pixel)} \\
 \hline
  & Before Refinement & Post Refinement\\
 \hline
 Dataset 1 & \textcolor{red}{51.642} & \textcolor{blue}{10.428}\\ \hline
 Dataset 2 & \textcolor{red}{54.245} & \textcolor{blue}{5.675}\\ 
 \hline
\end{tabular}}
\caption{\small{Root Mean Squared Re-projection Error calculated by measuring the reprojection error of the estimated distributed camera positions on each fisheye camera keyframe.}}
\label{table: reprojection-error}
\vspace{-1.75em}
\end{table}
One of the reasons behind the misalignment (in Figure \ref{fig: visual-residual}) can be limitations in estimating the spatial offset b/w fisheye camera and ArUco marker $T^{F}_{M}$ in Equation \ref{eqn: mocal-optimization} resulting from planar robot motion \cite{hecalib}. Since we cannot substantially improve $T^{F}_{M}$'s estimation, we refine the environment camera positions by minimizing reprojection error. We present the root mean squared reprojection error  (RMSE) over the entire robot trajectory for both the datasets in Table \ref{table: reprojection-error}, where we quantitatively show that the reprojection error post optimization is smaller. 

\subsection{Comparison with Lidar Odometry (LO) based method}
\label{sec: comparison-with-ifl}
\begin{figure*}[h]
        \centering
        \includegraphics[width=0.9\textwidth]{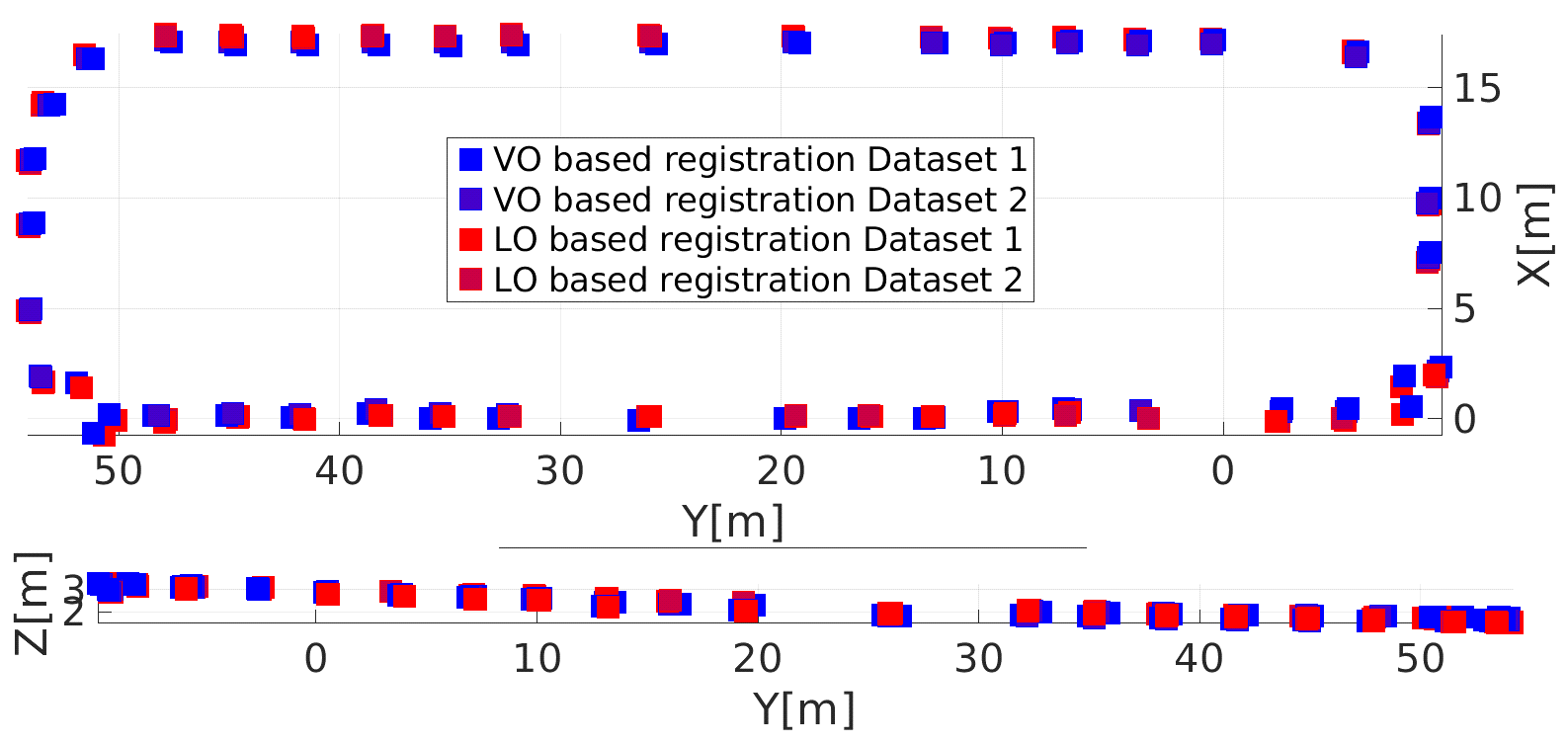}\label{fig: vo-lo-landmark}
        \caption{\small{Comparison of estimated env. camera poses between the VO and LO methods. Top and side view of env. camera positions estimated by VO and LO based approaches for dataset 1 $\&$ 2.}}
        \label{fig: vo-lo-landmark}
       \vspace{-1.25em}
\end{figure*}
In the absence of ground truth, we compare our method against the state of the art Direct LiDAR Odometry (DLO \cite{DLO}) to do LiDAR odometry (LO) and then find environmental camera poses. The LO based approach implements only Section \ref{sec: estimateposes-lookingdown} for the estimation process as two way detection is not possible when using a LiDAR. As LiDAR measures 3D points in metric units, it is not necessary to determine the metric scale of LO, but the spatial offset between the LiDAR and the ArUco marker is determined using motion based calibration \cite{MocalZachTaylor} method discussed in Section \ref{sec: mocal}.
Our goal is to discover if the VO based solution presented here gives comparable results. In order to compare the absolute environment camera poses we represent estimated variables in a common frame of reference (using a method similar to \cite{trajEval}). Visualization of the environment camera positions for both the approaches and both datasets is presented in Figure \ref{fig: vo-lo-landmark}. The RMS difference (RMSD) between corresponding environment cameras poses estimated using both the VO and LO based approaches for both the datasets are presented in Table \ref{table: rmse-absolute-rmse}, where we observe that the results from VO and LO are comparable, with the maximum rotation difference being 5.07$^{\circ}$ and maximum translation difference being 35.8 cm.


\begin{table}[h]
\centering
\resizebox{0.95\columnwidth}{!}{%
\begin{tabular}{| c | c |  c  | c || c |  c  | c |} 
\hline
\multicolumn{7}{|c|}{RMSD between corresponding environment cameras}\\
\hline & $\phi^\circ$ & $\theta^\circ$ & $\psi^\circ$ & X[m] & Y[m] & Z[m] \\
\hline
 $LO_1 \leftrightarrow VO_1$ & 0.9582  &  0.9220  &  4.0727 & 0.2819   & 0.3579 & 0.0947 \\
 \hline
 $LO_1 \leftrightarrow VO_2$ & 0.3604   & 1.3453  &  5.0665
 & 0.2281   & 0.1314  &  0.1104 \\
 \hline
 $LO_2 \leftrightarrow VO_1$ & 0.9768  &  1.3204  &  1.6038
 & 0.2662  &  0.3376  &  0.1029 \\
 \hline
 $LO_2 \leftrightarrow VO_2$ & 0.3777  &  0.7215  &  1.6997 & 0.2099 &   0.1433  &  0.1182 \\
 \hline
\end{tabular}}
\caption{\small{RMSD for estimated env. camera poses between VO and LO based approaches for both the datasets. Here $LO_i$ : environment cameras estimated from LO based approach in dataset $i$ $\&$ $VO_j$ : Corresponding environment cameras estimated from VO based approach in dataset $j$.}}
\label{table: rmse-absolute-rmse}
\vspace{-3em}
\end{table}

\section{DISCUSSION}
We present a method to register a large network of environment cameras using a calibration robot equipped with an inexpensive upward-facing fisheye camera and a backlit marker. We register $\sim$40 cameras in an perimeter of $\sim$150-155 m. We provide experimental results on two different datasets and compare our results with a LiDAR-based approach (DLO). In contrast to LiDAR-based approach, two-way sensing is possible with the visual approach, i.e. not only do the environment cameras detect the robot but the environment cameras are also detected in the robot's fisheye camera. This two-way sensing further constrains the optimization. The proposed VO solution is not only comparable to LO, but also allows us to self-verify registration by projecting the estimated environment camera position onto the fisheye camera image (Figures \ref{fig: visual-residual} $\&$ Figures \ref{fig: visual-residual-optimized}) which is not possible in the LO approach. Also, a camera is smaller, lighter, cheaper, and consumes less power, making its use possible with cheaper robotic platforms that can be used instead of the one used in this work. The registered environment cameras shall be used for downstream tasks like object detection, vehicle localization and navigation. The blob detection does not determine the exact environment camera optical centers, but our reprojections are within the pixels associated with the corresponding cameras, and as the dimensions of the environment cameras are known, we have an upper bound on the associated uncertainty.
\vspace{-1em}
\section{FUTURE WORK}
We plan to detect the environment cameras' corners to use its CAD model for estimating the optical center (by solving a PnP problem for each camera). Future work will also explore the usage of a drone with an upward facing camera instead of a wheeled calibration robot whose planar motion does not ensure full observability of the estimated parameters.


\bibliographystyle{ieeetr}
\bibliography{refs}


\end{document}